\documentclass[10pt,twocolumn]{article}
\usepackage[utf8]{inputenc}
\usepackage[T1]{fontenc}

\usepackage{amsmath,amssymb}
\usepackage{graphicx}
\usepackage{booktabs}
\usepackage{tabularx}
\usepackage{multirow}
\usepackage{subcaption}
\usepackage{longtable}
\usepackage{array}
\usepackage{adjustbox}
\usepackage{lscape}
\usepackage{enumitem}
\usepackage{xcolor}
\usepackage{hyperref}
\usepackage{caption}
\usepackage{geometry}

\usepackage{newunicodechar}
\newunicodechar{≡}{\equiv}

\title{\textbf{Clinical Knowledge Graph Construction and Evaluation with Multi-LLMs via Retrieval-Augmented Generation}}

\author{
Udiptaman Das$^{1}$ \and
Krishnasai B. Atmakuri$^{1}$ \and 
Duy Ho$^{2}$ \and 
Chi Lee$^{1}$ \and
Yugyung Lee$^{1}$ 
\\ 
$^{1}$University of Missouri--Kansas City, USA \\
$^{2}$California State University, Fullerton, USA \\
\texttt{\{ud3d4, bka2bg\}@umkc.edu duyho@fullerton.edu, 
\{leech,leeyu\}@umkc.edu}
}

\date{}  % No date for arXiv

\begin{document}
\maketitle

% =====================================================
% Abstract
% =====================================================
\begin{abstract}
Large language models (LLMs) offer new opportunities for constructing knowledge graphs (KGs) from unstructured clinical narratives. However, existing approaches often rely on structured inputs and lack robust validation of factual accuracy and semantic consistency—limitations that are especially problematic in oncology. We introduce an end-to-end framework for clinical KG construction and evaluation from free text using multi-agent prompting and a schema-constrained Retrieval-Augmented Generation (KG-RAG) strategy. Our pipeline integrates: (1) prompt-driven entity, attribute, and relation extraction; (2) entropy-based uncertainty scoring; (3) ontology-aligned RDF/OWL schema generation; and (4) multi-LLM consensus validation for hallucination detection and semantic refinement. Beyond static construction, the framework supports continuous refinement and self-supervised evaluation to iteratively improve graph quality. Applied to two oncology cohorts (PDAC and BRCA), our method produces interpretable, SPARQL-compatible, and clinically grounded knowledge graphs without gold-standard annotations, achieving consistent gains in precision, relevance, and ontology compliance over baseline methods.
\end{abstract}

% =====================================================
% Body starts here
% =====================================================

\section{Introduction}

Constructing accurate and clinically relevant knowledge graphs (KGs) from unstructured medical narratives is a foundational challenge in biomedical informatics. Clinical KGs enable explainable AI, decision support, and longitudinal patient modeling, yet traditional approaches remain limited. Rigid schemas like FHIR often lack semantic flexibility \cite{ayaz2021fast}, and while ontologies such as SNOMED CT \cite{snomed2022}, LOINC \cite{loinc2022}, and RxNorm \cite{rxnorm2022} offer standard terminologies, they struggle to capture the temporal, contextual, and inferential nuances needed for oncology and other complex domains.

Conventional rule-based or manual KG construction pipelines are brittle, difficult to scale, and ill-suited to the evolving language and structure of clinical narratives. In contrast, large language models (LLMs) have emerged as powerful tools for semantic parsing, relation discovery, and context-aware generation. Models such as \textit{Gemini 2.0 Flash} \cite{gemini2_2024}, \textit{GPT-4o} \cite{openai2024gpt4o}, and \textit{Grok 3} \cite{grok2024} show promise for automating KG construction directly from text. However, their outputs are prone to hallucinations, semantic drift, and factual inconsistency—issues especially critical in high-stakes domains like oncology \cite{rajpurkar2023evaluating, cao2023hallucination}.

Recent works such as \textit{CancerKG.ORG} \cite{gubanov2024cancerkg}, \textit{EMERGE} \cite{zhu2024emerge}, and \textit{CLR2G} \cite{xue2024clr2g} explore LLM integration with structured or multimodal sources, but lack generalizable pipelines for KG construction and validation directly from clinical text.

This paper introduces the first end-to-end framework for constructing and evaluating clinical knowledge graphs from free-text using multi-agent prompting and a graph-based Retrieval-Augmented Generation (KG-RAG) approach. Our pipeline supports continuous refinement and self-supervised evaluation, enabling both high-precision construction and dynamic graph improvement over time.

We leverage a multi-agent LLM pipeline combining the complementary strengths of:
\begin{itemize}[leftmargin=2em]
    \item \textit{Gemini 2.0 Flash} for schema-guided Entity–Attribute–Value (EAV) extraction;
    \item \textit{GPT-4o} for contextual enrichment, ontology alignment, and reflection-based refinement;
    \item \textit{Grok 3} for validation through contradiction testing and conservative filtering.
\end{itemize}

All triples are mapped to SNOMED CT, LOINC, RxNorm, GO, and ICD, and encoded in RDF/RDFS/OWL for semantic reasoning. Trust metrics are derived from model agreement and alignment with biomedical ontologies.
We demonstrate our method on 40 clinical oncology reports from the CORAL dataset \cite{sushil2024coral}, spanning PDAC and BRCA cohorts, and evaluate triple correctness, ontology coverage, relation diversity, and graph connectivity.

Our key contributions are as follows.
\begin{itemize}[leftmargin=2em]
    \item We propose a KG-RAG framework that integrates multi-agent prompting, LLM-based refinement, and continuous evaluation without gold-standard labels.
    \item We introduce a schema-constrained, FHIR-aligned EAV extraction module paired with ontology mapping and OWL-based encoding.
    \item We support inference over implicit, cross-sentence, and multi-attribute relations with robust semantic validation.
    \item We encode the final graphs with semantic web standards and introduce composite trust scoring mechanisms.
    \item We empirically validate our system on oncology narratives using metrics spanning factuality, semantic grounding, and relational completeness.
\end{itemize}

By uniting multi-LLM consensus, ontology grounding, and iterative refinement, our system offers a scalable, explainable, and verifiable approach for clinical KG construction—laying the foundation for real-time decision support and next-generation clinical AI.

\section{Related Work}

Recent advancements in hybrid knowledge graph–language model (KG–LLM) systems, retrieval-augmented generation (RAG), and contrastive learning have significantly influenced clinical informatics. However, many existing approaches rely on predefined schemas, curated corpora, or multimodal data. Our work addresses a critical gap: enabling schema-inductive, verifiable knowledge graph construction directly from unstructured clinical narratives.

\subsubsection*{\bf Schema-Constrained KG–LLM Hybrids}
{\it CancerKG.ORG}~\cite{gubanov2024cancerkg} integrates large language models (LLMs) with curated colorectal cancer knowledge using models such as GPT-4, LLaMA-2, and FLAN-T5, alongside structured meta-profiles and guardrailed RAG. Other frameworks like \textit{KnowGL}~\cite{liu2022knowgl}, \textit{KnowGPT}~\cite{zhang2024knowgpt}, and \textit{KALA}~\cite{liu2023kala} demonstrate the efficacy of knowledge graph alignment but depend on static ontologies. In contrast, our system facilitates dynamic schema evolution through automatic entity–attribute–value (EAV) extraction, probabilistic triple scoring, and iterative ontology grounding—eliminating the need for manual curation or fixed schemas.

\subsubsection*{\bf Multimodal RAG and Predictive Systems}

Systems such as \textit{EMERGE}~\cite{zhu2024emerge}, \textit{MedRAG}~\cite{zhao2025medrag}, and \textit{GatorTron-RAG}~\cite{yang2023gatortronrag} combine clinical notes, structured electronic health record (EHR) data, and biomedical graphs for predictive tasks. While effective for diagnosis or risk stratification, these models focus on classification rather than symbolic knowledge representation. Our approach emphasizes interpretable, symbolic triple construction using a fully text-based pipeline—removing dependencies on external modalities or fixed task objectives.

\subsubsection*{\bf Contrastive and Trust-Aware LLM Generation}

Trust-aware LLM generation has been explored through methods like \textit{Reflexion}~\cite{shinn2023reflexion} and \textit{TruthfulQA}~\cite{lin2022truthfulqa}, which employ self-reflection and hallucination mitigation techniques. Similarly, \textit{CLR2G}~\cite{xue2024clr2g} applies cross-modal contrastive learning for radiology report generation. While these approaches enhance semantic control, our framework advances this by introducing multi-agent triple scoring, entropy-aware trust filtering, and reflection-based validation across LLMs such as Gemini, GPT-4o, and Grok—focusing specifically on symbolic reasoning from clinical text.

\subsubsection*{\bf Structured Retrieval and Table-Centric Knowledge Graphs}

Earlier platforms including \textit{WebLens}~\cite{khan2020weblens}, \textit{Hybrid.JSON}~\cite{simmons2017hybridjson}, and \textit{COVIDKG}~\cite{kandibedala2023covidkg} introduced scalable structural retrieval and metadata modeling for heterogeneous clinical data tables. \textit{CancerKG} extends this direction using vertical and horizontal metadata modeling over structured cancer databases. Our work fundamentally differs in modality and scope: rather than relying on table alignment or attribute normalization, we construct evolving clinical knowledge graphs directly from free-text reports using prompt-based decomposition, probabilistic scoring, and ontology-backed reasoning.

In summary, while existing systems have laid essential groundwork in structured knowledge extraction and LLM-KG integration, our contribution is a self-evaluating, schema-flexible pipeline that autonomously discovers, validates, and encodes clinical knowledge from raw narratives. By removing rigid schema dependencies and leveraging multi-LLM agreement for factual trust, we enable scalable, interpretable, and clinically relevant knowledge graph construction that dynamically adapts to new domains and document types.

\section{Methodology}
\subsection{Overview of Multi-Agent KG Construction and Evaluation Pipeline}

We propose a multi-agent framework for constructing clinically verifiable and semantically interoperable knowledge graphs (KGs) directly from oncology narratives. The system orchestrates three state-of-the-art large language models—\textit{Gemini 2.0 Flash}, \textit{GPT-4.o}, and \textit{Grok 3}—each assigned a specialized role across five modular stages. This architecture is designed to produce structured, explainable, and ontology-aligned knowledge graphs from unstructured clinical text.

\begin{enumerate}[leftmargin=2em]
    \item \textbf{EAV Extraction (Gemini 2.0 Flash):} Using tailored prompts and FHIR-aware templates, \textit{Gemini 2.0 Flash} performs extraction of \textit{Entity–Attribute–Value} (EAV) triples from free-text narratives. Extracted entities are linked to canonical clinical resource types.

    \item \textbf{Ontology Mapping (Gemini 2.0 Flash):} Attributes and values are mapped to standard biomedical vocabularies such as \textit{SNOMED CT}, \textit{LOINC}, and \textit{RxNorm}. This step normalizes terminology and facilitates semantic consistency across clinical concepts.

    \item \textbf{Relation Discovery (Gemini 2.0 Flash + GPT-4.o):} Semantic relationships between entities and attributes are identified through prompt-driven extraction and refinement. \textit{Gemini 2.0 Flash} generates candidate links, while \textit{GPT-4.o} refines and validates relation types using contextual understanding and ontology alignment.

    \item \textbf{Semantic Web Encoding (RDF/RDFS/OWL):} All triples are encoded using Semantic Web standards such as RDF, RDFS, and OWL. This enables symbolic reasoning, knowledge graph querying via SPARQL, and integration with external ontology-based systems.

    \item \textbf{KG Validation (Gemini 2.0 Flash + GPT-4.o + Grok 3):} A composite trust function is applied to evaluate the reliability of each triple. The trust score incorporates:
    \begin{itemize}
        \item \textit{Self-consistency from Gemini 2.0 Flash} (via re-prompting),
        \item \textit{Semantic grounding from GPT-4.o} (evidence retrieval and consistency checks),
        \item \textit{Robustness verification from Grok 3} (counterfactual and adversarial assessment).
    \end{itemize}
\end{enumerate}

This pipeline supports schema-flexible, ontology-informed, and model-validated knowledge graph construction from raw clinical narratives. The modular architecture facilitates downstream reasoning, querying, and integration into clinical decision support systems while maintaining traceability and explainability of the extracted information.

\subsection{Stage 1: FHIR-Guided EAV Extraction}

The first stage of our pipeline focuses on extracting structured clinical knowledge in the form of \textit{Entity–Attribute–Value (EAV)} triples. This is achieved using schema-guided prompting with \textit{Gemini 2.0 Flash}, integrating syntactic cues from the narrative and semantic constraints from the \textit{FHIR (Fast Healthcare Interoperability Resources)} specification. Entities are concurrently typed using FHIR to ensure semantic consistency and interoperability.

\paragraph{\bf Triple Extraction.}
Let $\mathcal{D} = \{x_1, x_2, \ldots, x_N\}$ be the corpus of clinical narratives. For each $x_i \in \mathcal{D}$, we construct a structured prompt $\pi(x_i)$ tailored for \textit{Gemini 2.0 Flash}, which generates a candidate set of EAV triples as:

\vspace{-0.75em}
\[
\mathcal{T}_i = f_{\theta}(\pi(x_i)) = \left\{(e_j, a_j, v_j)\right\}_{j=1}^{k}, \quad \forall i \in [1, N]
\]
\vspace{-1em}

Each triple $(e_j, a_j, v_j)$ includes:
$e_j \in \mathcal{E}_{\text{FHIR}}$: a FHIR-typed entity (e.g., \texttt{Procedure}, \texttt{Observation}), 
$a_j$: a clinical attribute, and 
$v_j$: a narrative-grounded value.

\paragraph{\bf FHIR Typing Function.}
Entities are normalized via a typing function $\phi_{\text{FHIR}} : \mathcal{E} \rightarrow \mathcal{E}_{\text{FHIR}}$, ensuring alignment to standard resource types for downstream mapping.

\paragraph{\bf Entropy-Based Value Confidence.}
To evaluate confidence in value predictions, we compute token-level entropy for $v_j$ given its sub-token distribution $P(v_j) = \{p_1, \ldots, p_m\}$:

\vspace{-0.75em}
\[
H(v_j) = -\sum_{t=1}^{m} p_t \log p_t
\]
\vspace{-1em}

Values with $H(v_j) > \delta$ (threshold $\delta$) are flagged for further validation or multi-model filtering.

\paragraph{\bf Illustrative EAV Triples.}
Examples include:
\begin{quote}
\small
\texttt{(Procedure, performed\_by, SurgicalOncologist)} \\
\texttt{(Observation, hasLabResult, CA 19-9)} \\
\texttt{(HER2 Status, determines, Trastuzumab Eligibility)}
\end{quote}

These EAVs serve as the foundation for ontology mapping, relation discovery, and semantic web encoding in the subsequent stages.

\subsection{Stage 2: Ontology Mapping \& Schema Construction}

To enable semantic reasoning and interoperability, extracted EAV concepts are mapped to standardized biomedical ontologies via LLM-guided retrieval and similarity alignment. \textit{Gemini 2.0 Flash} orchestrates this step and produces OWL/RDFS-compliant schemas for graph construction.

\paragraph{Ontology Vocabulary.} 
We define the ontology set:
\[
\mathcal{O} = \{\text{SNOMED CT}, \text{LOINC}, \text{RxNorm}, \text{ICD}, \text{GO}\}
\]
Each $\mathcal{O}_i$ contributes domain-specific concepts, e.g., SNOMED: \texttt{Weight Loss}, LOINC: \texttt{CA 19-9}, RxNorm: \texttt{FOLFIRINOX}.

\paragraph{Concept Mapping.} 
Given raw terms $\mathcal{C}_{\text{raw}} = \{c_1, \ldots, c_M\}$, define a mapping:
\[
\mu: \mathcal{C}_{\text{raw}} \rightarrow \mathcal{C}_{\text{mapped}} \subseteq \bigcup_i \mathcal{O}_i
\]
Scoring uses lexical and semantic similarity:
\[
\text{Score}(c_i, o_j) = \alpha \cdot \text{sim}_{\text{lex}} + \beta \cdot \text{sim}_{\text{sem}}, \quad \alpha + \beta = 1
\]

\paragraph{Schema Construction.}
Mapped concepts are encoded in OWL/RDFS:
\vspace{-0.5em}
\begin{itemize}[leftmargin=1.5em, itemsep=0.3em]
    \item \textbf{Class Typing:} If $o_j \in \mathcal{O}_{\text{SNOMED}}$, declare $o_j$ as an OWL \texttt{Class}.
    \item \textbf{Property Semantics:}
    \begin{align*}
        \texttt{hasLabResult} &\sqsubseteq \texttt{ObjectProperty}, \\
        \texttt{domain(hasLabResult)} &= \texttt{Observation}, \\
        \texttt{range(hasLabResult)} &= \texttt{LabTest}
    \end{align*}
    \item \textbf{TBox Inclusion:} \texttt{ElevatedCA19\_9} $\sqsubseteq$ \texttt{AbnormalTumorMarker}.
\end{itemize}

\paragraph{Persistent URIs.} 
Each concept receives a resolvable URI, e.g.,
\texttt{http://snomed.info/id/267036007} for “Weight Loss.”

\paragraph{Outcome.} 
This stage yields an ontology-aligned schema that supports OWL reasoning, SPARQL queries, and Linked Data integration, while ensuring formal consistency and semantic traceability.

\subsection{Stage 3: Relation Discovery}

To move beyond isolated EAV triples, this stage enriches the knowledge graph with typed relations capturing diagnostic reasoning, temporal dependencies, and treatment logic. Structured prompting and multi-agent validation (Gemini 2.0 Flash, GPT-4.o, Grok 3) are used to discover and filter semantic relations.

\textbf{Relation Typing.} Each relation $r_i \in \mathcal{R}$ is classified as:
\[
r_i \in 
\begin{cases}
\mathcal{R}_{EE} & \text{(Entity–Entity)} \\
\mathcal{R}_{EA} & \text{(Entity–Attribute)} \\
\mathcal{R}_{AA} & \text{(Attribute–Attribute)}
\end{cases}
\]
Examples include: $\mathcal{R}_{EE}$: \texttt{Biopsy $\rightarrow$ confirms $\rightarrow$ TumorType}, $\mathcal{R}_{EA}$: \texttt{CT Scan $\rightarrow$ visualizes $\rightarrow$ Pancreatic Mass}, $\mathcal{R}_{AA}$: \texttt{HER2 Status $\rightarrow$ determines $\rightarrow$ Trastuzumab Eligibility}.

\textbf{Candidate Generation.} Given narrative $x$, candidate relations $\mathcal{T}_{\text{rel}} = \{(h_i, p_i, t_i)\}$ are generated via Gemini: 
\[
\mathcal{T}^{\text{gen}}_{\text{rel}} = f_{\text{Gemini}}(\pi_{\text{rel}}(x))
\]
with $h_i, t_i \in \mathcal{E} \cup \mathcal{A}$ and $p_i \in \mathcal{V}_{\text{verb}}$.

\textbf{Semantic Validation.} Each triple $\tau_i$ is scored by GPT-4.o using contextual inference:
\[
J(\tau_i) = f_{\text{GPT-4.o}}(\pi_{\text{judge}}(\tau_i, x)) \in [0,1]
\]

\textbf{Adversarial Filtering.} Grok 3 perturbs each relation and flags contradictions:
\[
\xi(\tau_i) = \frac{|\{\tau_i' \in \mathcal{A}(\tau_i)\ |\ \texttt{contradictory}\}|}{|\mathcal{A}(\tau_i)|}
\]

\textbf{Final Set.} The accepted relation set filters for high plausibility and low contradiction:
\[
\mathcal{T}_{\text{rel}}^{\text{trusted}} = \left\{\tau_i \mid J(\tau_i) > \delta,\ \xi(\tau_i) \leq \epsilon \right\}
\]

\textbf{Outcome.} This step yields a validated set of typed relations interlinking FHIR-grounded concepts. These enhance the inferential depth of the KG, supporting diagnostic, prognostic, and treatment-oriented reasoning.

\subsection{Stage 4: Semantic Graph Encoding}

To enable reasoning and system interoperability, validated triples $\tau_i = (s_i, p_i, o_i)$ are encoded using Semantic Web standards: RDF, RDFS, OWL, and SWRL.

\textbf{RDF Triples.} Each assertion is modeled as: 
$(s_i, p_i, o_i) \in \mathcal{G}_{\text{RDF}}$, 
where $s_i, o_i \in \mathcal{U} \cup \mathcal{L}$ and $p_i \in \mathcal{U}$; 
$\mathcal{U}$ denotes URIs and $\mathcal{L}$ literals.

\textbf{RDFS Constraints.} Predicates include domain/range typing: 
$\text{domain}(p_i) = C_s$, $\text{range}(p_i) = C_o$, 
with type assertions:
$\texttt{rdf:type}(s_i, C_s) \land \texttt{rdf:type}(o_i, C_o)$.

\textbf{OWL Semantics.} Logical rules include: 
Subclass: $A \sqsubseteq B$, 
Equivalence: $A \equiv B$, 
Restriction: 
$\texttt{Biopsy} \sqcap \exists\,\texttt{hasOutcome}.\texttt{Malignant} \sqsubseteq \texttt{PositiveFinding}$.

\textbf{SPARQL Query.} Query for high Ki-67 index:
\begin{verbatim}
PREFIX kg: <http://example.org/kg#>
SELECT ?p WHERE {
  ?p kg:hasAttribute ?a .
  ?a rdf:type kg:Ki67_Index .
  ?a kg:indicates ?v .
  FILTER(?v > 20)
}
\end{verbatim}

\textbf{SWRL Rule.} Example rule for identifying high-risk PDAC patients:
\[
\begin{array}{l}
\texttt{Patient}(p) \land \texttt{hasAttribute}(p, ca) \land \texttt{CA19\_9}(ca) \land \\
\quad \texttt{indicates}(ca, v_1) \land \texttt{greaterThan}(v_1, 1000) \land \\
\quad \texttt{hasAttribute}(p, w) \land \texttt{WeightLoss}(w) \land \\
\quad \texttt{indicates}(w, v_2) \land \texttt{greaterThan}(v_2, 10) \\
\Rightarrow \texttt{HighRiskPatient}(p)
\end{array}
\]

\textbf{Outcome.} This stage produces an ontology-compliant semantic graph supporting RDFS validation, OWL/SWRL inference, and SPARQL querying—ideal for clinical integration and semantic analytics.

\subsection{Stage 5: Multi-LLM Trust Validation}

\begin{table*}[ht!]
\centering
\footnotesize
\caption{Three-Stage Matching Pipeline for EAV Triple Validation}
\vspace{-0.1in}
\label{tab:eav_matching_pipeline}
\begin{tabular}{|p{3.5cm}|p{5cm}|p{7.5cm}|}
\hline
\textbf{Technique} & \textbf{Formal Rule} & \textbf{Illustrative Example} \\
\hline

\multicolumn{3}{|c|}{\textbf{Stage 1: Baseline Matching Techniques}} \\
\hline
Case-Sensitive Matching & $v_i \in \mathcal{T}_{\text{exact}}(d_j)$ & “FOLFIRINOX” matched exactly in raw text \\
Regex Matching & \texttt{regex("1\.2\ ?mg/dL")} & \texttt{bilirubin: 1.2 mg/dL} matched as value \\
Fuzzy String Matching & $\texttt{fuzz.ratio}(v_i, t_k) > \tau$ & “FOLFRINOX” $\rightarrow$ “FOLFIRINOX” \\
N-Gram Phrase Matching & $\exists\, g \subset d_j:\ \text{sim}(g, v_i) > \gamma$ & “2–3 episodes/day” matched to vomiting frequency \\
Boolean Inference & “denies smoking” $\Rightarrow$ \texttt{smoking: false} & Negation mapped to absent finding \\
\hline

\multicolumn{3}{|c|}{\textbf{Stage 2: Heuristic Augmentation}} \\
\hline
Case-Insensitive Matching & \texttt{lower(v\_i) = lower(t\_k)} & “Male” == “male” == “MALE” \\
Custom Negation Detection & “denies chills” $\rightarrow$ \texttt{observation\_chills: absent} & Handcrafted negation patterns \\
spaCy Lemmatization & $\texttt{lemma}(a_i) = t_k$ & “diaphoretic” → “diaphoresis” \\
Synonym Mapping & $a_i \in \Sigma_{\text{syn}}$ & ``dyspneic'' $\leftrightarrow$ ``dyspnea'' \\
\hline

\multicolumn{3}{|c|}{\textbf{Stage 3: Specialized Resolution}} \\
\hline
Sentence-Level Negation & $\text{dep}(t_k) = \text{neg}$ in sentence($v_i$) & “denies pruritus” → \texttt{absent} \\
Typo Correction & $\texttt{fuzz.ratio}(\texttt{deneid}, \texttt{denied}) > 95$ & Handles transcription errors \\
Explicit Fixes & “smking” → “smoking” & Dictionary-based recovery of high-impact terms \\
\hline
\end{tabular}
\end{table*}

\begin{table*}[ht!]
\centering
\footnotesize
\caption{Relation-Level Validation Techniques and Examples}
\vspace{-0.1in}
\label{tab:relation_validation}
\begin{tabular}{|p{3.5cm}|p{5cm}|p{7.5cm}|}
\hline
\textbf{Technique} & \textbf{Formal Rule or Criterion} & \textbf{Illustrative Example} \\
\hline

Evidence Alignment (Entailment) & $\text{Sim}_{\text{entail}}(\tau, e_k)$ for $e_k \in \mathcal{E}_\tau$ & “HER2 is overexpressed” entails \texttt{HER2 $\rightarrow$ determines $\rightarrow$ Trastuzumab Eligibility} \\
\hline
LLM Reflective Scoring & $J(\tau) \in [0, 1]$ from GPT-4o verifier prompt & GPT-4o returns 0.92 plausibility for therapy eligibility triple \\
\hline
Self-Consistency across Prompts & $C(\tau) = \frac{1}{n} \sum \mathbb{I}[\tau \in \mathcal{T}^{(i)}]$ & Relation appears in 4/5 prompt variants → $C = 0.80$ \\
\hline
Domain-Range Schema Check & $\text{type}(s) \in \text{domain}(p)$; $\text{type}(o) \in \text{range}(p)$ & \texttt{Tumor → hasMarker → HER2} invalid if HER2 is not range of hasMarker \\
\hline
Redundancy Score & $\cos(\tau_i, \tau_j) > \gamma$ in embedding space & “confirms” and “verifies” flagged as duplicates \\
\hline
Semantic Clustering for Gaps & Use UMAP clustering over embeddings & Reveals that “initiated” and “started” lack edge alignment \\
\hline
\end{tabular}
\end{table*}

To ensure the reliability and semantic precision of the constructed clinical knowledge graph, we implement a multi-layered validation framework. This framework consists of two complementary tiers: (1) evaluating the fidelity of Entity–Attribute–Value (EAV) triples and (2) verifying semantic relation triples. Both tiers are supported by a multi-agent setup involving \textit{Gemini 2.0 Flash}, \textit{GPT-4.o}, and \textit{Grok 3}, which collaboratively perform grounding verification, reflective reasoning, and adversarial testing.

\subsubsection*{\bf EAV Validation: Grounding and Fidelity Assessment}

Each EAV triple $(e, a, v)$ is validated across multiple quality dimensions, including textual grounding, correctness, hallucination risk, recoverability, and prompt-based consistency.

\paragraph{\bf EAV Validation Metrics.}
\begin{itemize}[leftmargin=2em]
    \item \textit{Coverage:} Whether the attribute $a$ and value $v$ are directly supported by the source text.
    \item \textit{Correctness Rate (CR):} Proportion of grounded EAV triples verified as correct.
    \item \textit{Hallucination Rate (HR):} Fraction of generated triples lacking explicit textual support.
    \item \textit{Rescue Rate (RR):} Proportion of hallucinated triples that become valid after normalization or lexico-semantic heuristics.
    \item \textit{Self-Consistency:} Agreement across multiple prompting strategies applied to the same source.
\end{itemize}

\subsubsection*{\bf Relation Validation: Semantic Integrity and Clinical Utility}

Each relation triple $(s, p, o)$ is evaluated on criteria that assess its semantic validity, ontological alignment, clinical relevance, and informativeness within the graph.

\paragraph{\bf Relation Evaluation Criteria.}
\begin{itemize}[leftmargin=2em]
    \item \textit{Semantic Validity:} Verified via entailment and alignment with external evidence or ontologies.
    \item \textit{Schema Compliance:} Ensures each predicate respects RDF domain and range constraints.
    \item \textit{Clinical Usefulness:} Prioritizes relations with diagnostic, prognostic, or therapeutic significance.
    \item \textit{Structural Role and Frequency:} Measures relational salience based on graph centrality and recurrence.
    \item \textit{Redundancy and Gaps:} Detects paraphrased or missing links using embedding-based similarity.
    \item \textit{Multi-LLM Agreement:} Confidence derived from consensus across \textit{Gemini 2.0}, \textit{GPT-4.o}, and \textit{Grok 3} outputs.
\end{itemize}

\subsubsection*{\bf Unified Trust Scoring}

A composite trust score $T(\tau)$ is assigned to each triple $\tau = (s, p, o)$ or $(e, a, v)$:
\[
T(\tau) = \lambda_1 R(\tau) + \lambda_2 C(\tau) + \lambda_3 J(\tau), \quad \sum \lambda_i = 1
\]
Where:
\begin{itemize}[leftmargin=2em]
    \item $R(\tau)$: Evidence alignment score from entailment or retrieval.
    \item $C(\tau)$: Self-consistency across prompt variants.
    \item $J(\tau)$: Reflective plausibility score from model-based judgment.
\end{itemize}

Only triples satisfying $T(\tau) \geq \delta_T$ (e.g., $\delta_T = 0.65$) are retained in the final knowledge graph.

\subsubsection*{\bf Final Graph Validation and Filtering}
The final graph $\mathcal{G}_{\text{final}}$ is further validated through graph-level filters:
\begin{itemize}[leftmargin=2em]
    \item \textit{Ontology Compliance:} Ensures that all relations adhere to predefined schema constraints.
    \item \textit{Redundancy Elimination:} Removes duplicate or semantically equivalent edges.
    \item \textit{Clinical Coverage:} Emphasizes inclusion of medically relevant concepts (e.g., NCCN-aligned markers, therapeutic eligibility).
\end{itemize}

\paragraph{\bf Outcome.}
This multi-agent, multi-criteria validation framework ensures that the final knowledge graph is clinically grounded, semantically coherent, and structurally optimized. The result is a trustworthy KG suitable for deployment in applications such as explainable AI, cohort identification, treatment planning, and clinical decision support.

\section{Results and Evaluation}

\begin{figure*}[ht!]
\vspace{-0.1in}
\centering
\includegraphics[width=1\linewidth]{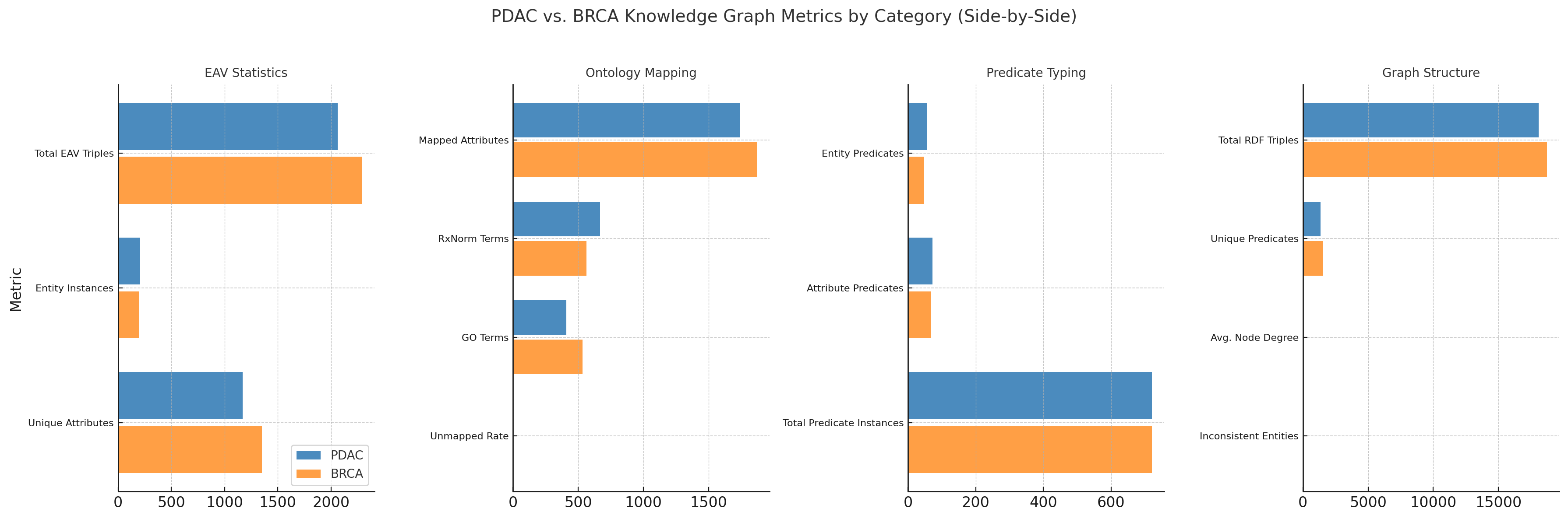}
\vspace{-0.1in}
\caption{PDAC vs. BRCA Knowledge Graph Metrics by EAV Statistics, Ontology Mapping, Predicate Typing, and Graph Structure. BRCA emphasizes molecular attributes; PDAC reflects procedural diversity.}

\label{fig:kg_summary_plot}
\vspace{-0.1in}
\end{figure*}

\begin{table*}[ht!]
\centering
\footnotesize
\setlength{\tabcolsep}{3pt}
\renewcommand{\arraystretch}{1.15}
\caption{Consolidated knowledge graph metrics with examples (PDAC vs.\ BRCA).}
\vspace{-0.1in}
\label{tab:kg_summary_all}

\begin{tabularx}{\textwidth}{@{}p{2.8cm} p{3.3cm} c c X@{}}
\toprule
\textbf{Metric Category} & \textbf{Metric} & \textbf{PDAC} & \textbf{BRCA} & \textbf{Example / Description} \\
\midrule

\multirow{3}{*}{\textbf{EAV Statistics}}
& Total EAV triples & 2,062 & 2,293 & Structured triples such as \texttt{(Observation, hasLabResult, CA\_19\mbox{-}9)}. \\
& Entity instances & 209 & 195 & Entity types like \texttt{Procedure}, \texttt{Observation}, \texttt{Condition}. \\
& Unique attributes & 1,172 & 1,353 & Fine-grained attributes like \texttt{Weight\_Loss}, \texttt{HER2\_Status}. \\
\midrule

\multirow{4}{*}{\textbf{Ontology Mapping}}
& Mapped attributes & 1,738 & 1,873 & Attributes linked to SNOMED CT, RxNorm, and LOINC. \\
& RxNorm terms & 668 & 562 & \texttt{FOLFIRINOX}, \texttt{Tamoxifen}, chemotherapy agents. \\
& GO terms & 409 & 533 & Genomic markers like \texttt{HER2}, \texttt{BRCA1}, \texttt{EGFR}. \\
& Unmapped rate & 0.69\% & 0.85\% & Mostly generic strings, typos, or shorthand entries. \\
\midrule

\multirow{3}{*}{\textbf{Predicate Typing}}
& Entity predicates & 56 & 46 & Action relations: \texttt{(Biopsy, confirms, TumorType)}. \\
& Attribute predicates & 72 & 69 & Diagnostic relations: \texttt{(HER2Status, determines, TherapyEligibility)}. \\
& Total predicate instances & 721 & 721 & Total predicate uses across the graph. \\
\midrule

\multirow{4}{*}{\textbf{Graph Structure}}
& Total RDF triples & 18,097 & 18,732 & RDF-format atomic facts encoding EAV and relations. \\
& Unique predicates & 1,346 & 1,520 & Vocabulary such as \texttt{hasLabResult}, \texttt{performedBy}. \\
& Avg.\ node degree & 5.73 & 5.74 & Graph density and information connectivity. \\
& Inconsistent entities & 25 & 41 & Domain--range issues caught during OWL validation. \\
\bottomrule
\end{tabularx}

\vspace{-0.1in}
\end{table*}

\subsection{Experimental Setup}

We evaluate our clinical knowledge graph (KG) framework using 40 expert-annotated oncology reports from the \textit{CORAL (Clinical Oncology Reports to Advance Language Models)} dataset. The reports span two cancer types—\textit{Pancreatic Ductal Adenocarcinoma (PDAC)} and \textit{Breast Cancer (BRCA)}—which differ significantly in diagnostic focus and treatment documentation.

Our primary objectives are to assess the grounding and accuracy of extracted \textit{Entity–Attribute–Value (EAV)} triples and to evaluate the structural and semantic integrity of the resulting knowledge graphs.
The KG construction pipeline integrates four core components. First, \textit{Gemini 2.0 Flash} is used to extract FHIR-aligned EAV triples from unstructured narratives. Second, \textit{GPT-4.o}, \textit{Grok 3}, and Gemini collaborate to validate triples—filtering hallucinations and resolving inconsistencies. Third, extracted terms are normalized to standardized vocabularies, including SNOMED CT, RxNorm, LOINC, GO, and ICD. Finally, validated triples are encoded into RDF/RDFS/OWL to support SPARQL and SWRL-based rule reasoning.

Each of the 40 clinical reports—20 PDAC and 20 BRCA—contains a narrative file (\texttt{.txt}), expert annotations (\texttt{.ann}), and model-generated outputs (\texttt{.json}). The average report is approximately 145,000 tokens, with some extending to 180,000 tokens in more complex cases.
The three LLMs play distinct but complementary roles. \textit{Gemini 2.0 Flash} acts as the primary EAV extractor, aligning outputs to FHIR standards and providing confidence scores. \textit{GPT-4.o} validates predicate plausibility and assigns reflection-based trust levels. \textit{Grok 3} performs adversarial testing to detect redundancy and semantic inconsistencies.
This multi-agent framework enables robust, interpretable, and ontology-compliant knowledge graph construction directly from raw clinical narratives.

\subsection{Knowledge Graph Construction Results}

Figure~\ref{fig:kg_summary_plot} and Table~\ref{tab:kg_summary_all} summarize the semantic and structural features of knowledge graphs constructed from PDAC and BRCA oncology reports. The visual and tabular comparisons reflect all key pipeline stages—EAV extraction, ontology mapping, predicate typing, and semantic modeling—and highlight domain-specific characteristics across cancer types. Together, they demonstrate the fidelity, flexibility, and interpretability of our multi-agent LLM-based KG construction approach.

\vspace{0.5em}
\noindent\textbf{Step 1: FHIR-Aligned EAV Extraction.}
BRCA records yielded more EAV triples (2,293 vs. 2,062) and a broader attribute set (1,353 vs. 1,172), reflecting molecular precision. Examples:
%\vspace{-0.4em}
\begin{itemize}[leftmargin=2em, itemsep=0em]
    \item \texttt{HER2 Status}$\rightarrow$\texttt{determines}$\rightarrow$\texttt{Trastuzumab Eligibility}
    \item \texttt{Ki-67 Index}$\rightarrow$\texttt{indicates}$\rightarrow$ \texttt{High Proliferation Rate}
\end{itemize}
\vspace{-0.4em}
PDAC emphasized diagnostic and procedural concepts:
\vspace{-0.4em}
\begin{itemize}[leftmargin=2em, itemsep=0em]
    \item \texttt{CT Scan}$\rightarrow$ \texttt{visualizes}$\rightarrow$ \texttt{Pancreatic Mass}
    \item \texttt{Surgical Resection}$\rightarrow$ \texttt{treats}$\rightarrow$ \texttt{Primary Tumor}
\end{itemize}

\vspace{0.5em}
\noindent\textbf{Step 2: Ontology Mapping.}
BRCA had more mapped attributes (1,873 vs. 1,738) and GO terms (533 vs. 409), consistent with genomics-rich narratives. PDAC showed stronger RxNorm alignment (668 vs. 562), reflecting treatment-oriented notes. Both cohorts had <1\% unmapped rate.

\vspace{0.5em}
\noindent\textbf{Step 3: Predicate Typing.}
Both KGs included 721 predicate instances, with PDAC using a broader predicate range (56 entity-level, 72 attribute-level) than BRCA (46, 69). PDAC predicates emphasized procedural workflows, BRCA prioritized stratification and therapeutic relevance.

\vspace{0.5em}
\noindent\textbf{Step 4: Graph Structure and Semantic Modeling.}
Both graphs showed dense connectivity (avg. node degree $\sim$5.7) and rich vocabularies (PDAC: 1,346 predicates; BRCA: 1,520). OWL validation found more inconsistencies in BRCA (41) than PDAC (25), often from genomics-based domain-range mismatches. Triples were encoded in RDF/RDFS/OWL, supporting SPARQL/SWRL:
%\vspace{-0.5em}
\[
\texttt{Biopsy} \sqcap \texttt{hasOutcome.Malignant} \sqsubseteq \texttt{PositiveFinding}
\]

\vspace{0.5em}
\noindent\textbf{Clinical Interpretability.}
Actionable triples like \texttt{(Biopsy, confirms, TumorType)} and \texttt{(HER2 Status, determines, Trastuzumab Eligibility)} demonstrate support for clinical decision-making and cohort stratification. Our framework—Gemini 2.0 Flash for generation, GPT-4o for validation, Grok 3 for semantic robustness—transforms unstructured narratives into queryable, ontology-aligned knowledge graphs.

\subsection{Evaluation of  Knowledge Graphs}

We evaluate the quality of LLM-generated clinical knowledge graphs (KGs) using a two-tiered framework: (1) analysis of Entity–Attribute–Value (EAV) triples for textual and semantic fidelity, and (2) structural and relation-level validation of the graph itself. Evaluation spans two cancer cohorts—PDAC and BRCA—derived from the CORAL dataset.

\subsubsection{\bf Entity–Attribute–Value (EAV) Evaluation}

\vspace{-0.05in}
\subsubsection*{\bf Evaluation of EAV Extraction Across PDAC and BRCA Cohorts}

We conducted a comparative evaluation of entity-attribute-value (EAV) extraction across two cancer cohorts: Pancreatic Ductal Adenocarcinoma (PDAC, Patients 0--19) and Breast Cancer (BRCA, Patients 20--39). Key metrics include raw text and attribute coverage, correctness, hallucinations, and error rates, summarized in Table~\ref{tab:eav_summary_detailed}.

\begin{table*}[ht]
\centering
\footnotesize
%\vspace{-0.1in}
\caption{EAV Quality Metrics Across Cancer Cohorts}
\vspace{-0.1in}
\label{tab:eav_summary_detailed}
\begin{tabular}{p{4cm}|c|c|c|c}
\hline
\multirow{2}{*}{\textbf{Metric}} & \multicolumn{2}{c|}{\textbf{PDAC (P\# 0--19)}} & \multicolumn{2}{c}{\textbf{BRCA (P\# 20--39)}} \\
\cline{2-5}
& \textbf{Avg / Rate} & \textbf{Total} & \textbf{Avg / Rate} & \textbf{Total} \\
\hline
\multicolumn{5}{l}{\textbf{Text \& Attribute Coverage}} \\
\hline
Raw Text Coverage (\%)              & 29.74\% & --   & 37.97\% & --   \\
Attribute Coverage (\%)            & 99.83\% & --   & 99.83\% & --   \\
Total Attributes Extracted         & --      & 2,028 & --      & 2,099 \\
\hline
\multicolumn{5}{l}{\textbf{Correctness \& Error Metrics}} \\
\hline
Correct Attributes (\%)            & 73.22\% & 1,514 & 72.58\% & 1,524 \\
Incorrect Attr. (Error\#) & 25.45   & 509  & 28.25   & 565  \\
Errors (Attr.) (\%)         & 25.10\% & --   & 26.91\% & --   \\
\hline
\multicolumn{5}{l}{\textbf{Hallucination \& Consistency}} \\
\hline
Hallucinated Attr. (Total)    & 0.20    & 4    & 0.15    & 3    \\
\hline
\end{tabular}
%\vspace{-0.1in}
\end{table*}

BRCA patients showed higher average raw text coverage (37.97\%) than PDAC (29.74\%), suggesting denser or more structured narratives in breast cancer reports. PDAC achieved slightly better extraction precision, with a lower incorrect attribute rate (25.45 vs. 28.25) and a smaller overall error rate (25.10\% vs. 26.91\%). Both cohorts demonstrated excellent attribute coverage (99.83\%), indicating consistent LLM performance across cancer types. Hallucination rates were minimal—fewer than one per five patients—validating the reliability of prompt-based extraction.

These results highlight a trade-off between recall and precision. BRCA’s richer documentation may increase coverage but introduces more noise, while PDAC yields fewer attributes with higher correctness. This underscores the need for cohort-specific fine-tuning or filtering in future extraction pipelines.

\subsubsection*{\bf Attribute-Level Comparison}

\begin{figure*}[ht]
%\vspace{-0.1in}
\centering
\includegraphics[width=1\linewidth]{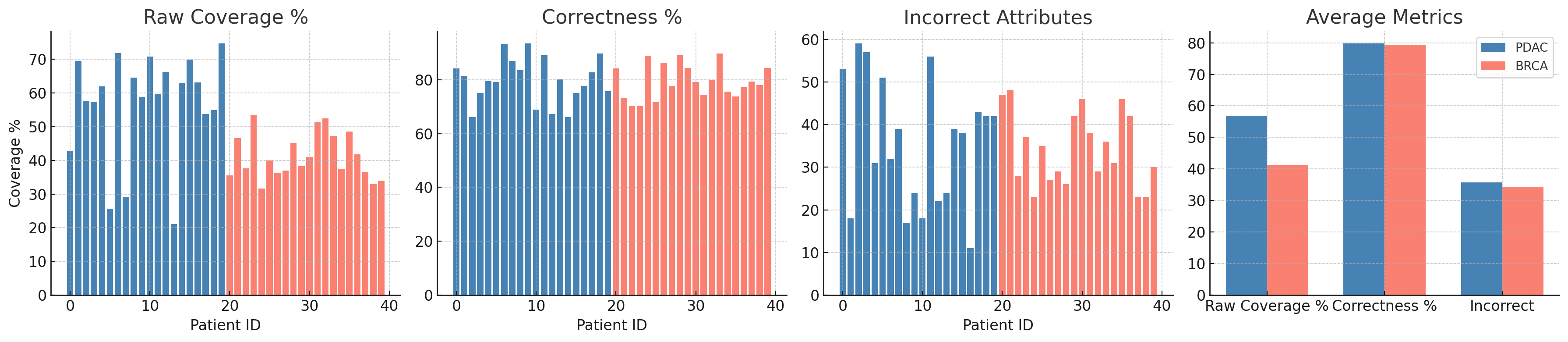}
\vspace{-0.1in}
\caption{Comparison of attribute-level metrics for PDAC and BRCA cohorts. From left to right: raw coverage percentage, correctness percentage, incorrect attribute counts, and overall averages.}
\label{fig:pdac_brca_attribute_eval}
%\vspace{-0.1in}
\end{figure*}

Figure~\ref{fig:pdac_brca_attribute_eval} presents a comparison of raw coverage, correctness, and error rates across PDAC and BRCA cohorts. PDAC patients show stable correctness with fewer attribute-level errors, whereas BRCA patients demonstrate greater variability, most notably among Patients 30 and 31, who exhibit high attribute volume but reduced correctness.

Both cohorts demonstrate strong EAV extraction. PDAC emphasizes higher correctness and reduced noise, while BRCA contributes richer attribute diversity. These patterns highlight the importance of tailoring extraction strategies to cohort-specific narrative styles, especially for oncology KGs and adaptive clinical decision support.

\subsubsection{\bf KG Structural and Relation-Level Validation}

%\vspace{-0.1in}
\subsubsection*{\bf Multi-LLM Consensus Evaluation of EAV Triples}

To ensure semantic fidelity and factual grounding of extracted Entity–Attribute–Value (EAV) triples, we adopt a consensus-driven validation framework using three complementary LLMs: \textit{Gemini 2.0 Flash} (schema-aware extractor), \textit{Grok 3} (evidence-based validator), and \textit{GPT-4o} (semantic generalizer). Triples are assessed across three dimensions: (1) \textit{Factuality}—explicit grounding in source text or biomedical ontologies, (2) \textit{Plausibility}—semantically inferable but implicit triples, and (3) \textit{Correction}—revision of ambiguous or hallucinated content.

Each model contributes distinct strengths: Grok filters unsupported relations (e.g., \texttt{hematuria\_etiology $\rightarrow$ diagnosis}); GPT-4o proposes contextually inferred alternatives (e.g., \texttt{creatinine $\rightarrow$ kidney\_function}); Gemini provides precise, schema-aligned outputs.

\begin{figure*}[ht]
    %\vspace{-0.1in}
    \centering
    \includegraphics[width=1\linewidth]{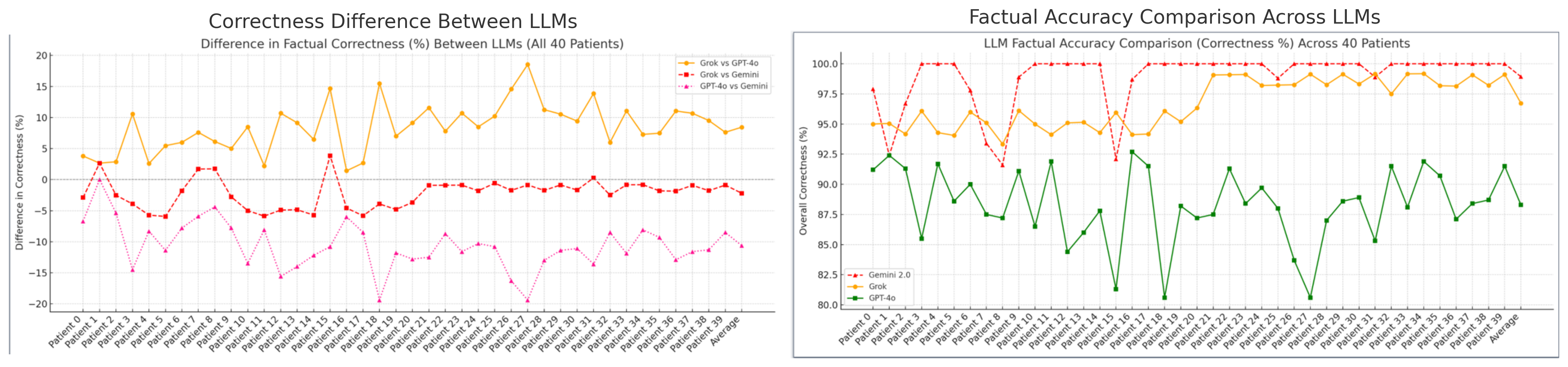}
    \vspace{-0.1in}
    \caption{Top: Per-patient factual correctness of EAV triples across three LLMs. Bottom: Pairwise correctness differences and Pearson correlation trends across 40 patients.}
    \label{fig:factual_accuracy_combined}
   % \vspace{-0.1in}
\end{figure*}

Figure~\ref{fig:factual_accuracy_combined} shows Gemini achieves the highest factual accuracy (98–100\%), followed by Grok (94–96\%) and GPT-4o (85–92\%). Pearson correlations indicate strong alignment between Gemini and Grok ($r = 0.88$), with GPT-4o moderately correlated with Gemini ($r = 0.71$) and Grok ($r = 0.69$), consistent with its broader semantic scope.

In the absence of gold-standard annotations, we accept triples validated by at least two models. Disagreements are resolved by prioritizing ontology-compliant alternatives (e.g., SNOMED CT, RxNorm), with uncertain triples flagged for reflection and reranking.
Together, the precision of Gemini, the filtering strength of Grok, and the contextual breadth of GPT-4o enable robust, unsupervised validation of clinical KGs. Correlation trends reinforce their complementary roles in accurate, ontology-aligned graph construction—without the need for human-labeled supervision.

%\vspace{-0.1in}
\subsubsection*{\bf Statistical Analysis of Relation Diversity and Source Coverage}

\begin{figure*}[ht]
\vspace{-0.1in}
\centering
\begin{subfigure}{0.32\textwidth}
    \includegraphics[width=\linewidth]{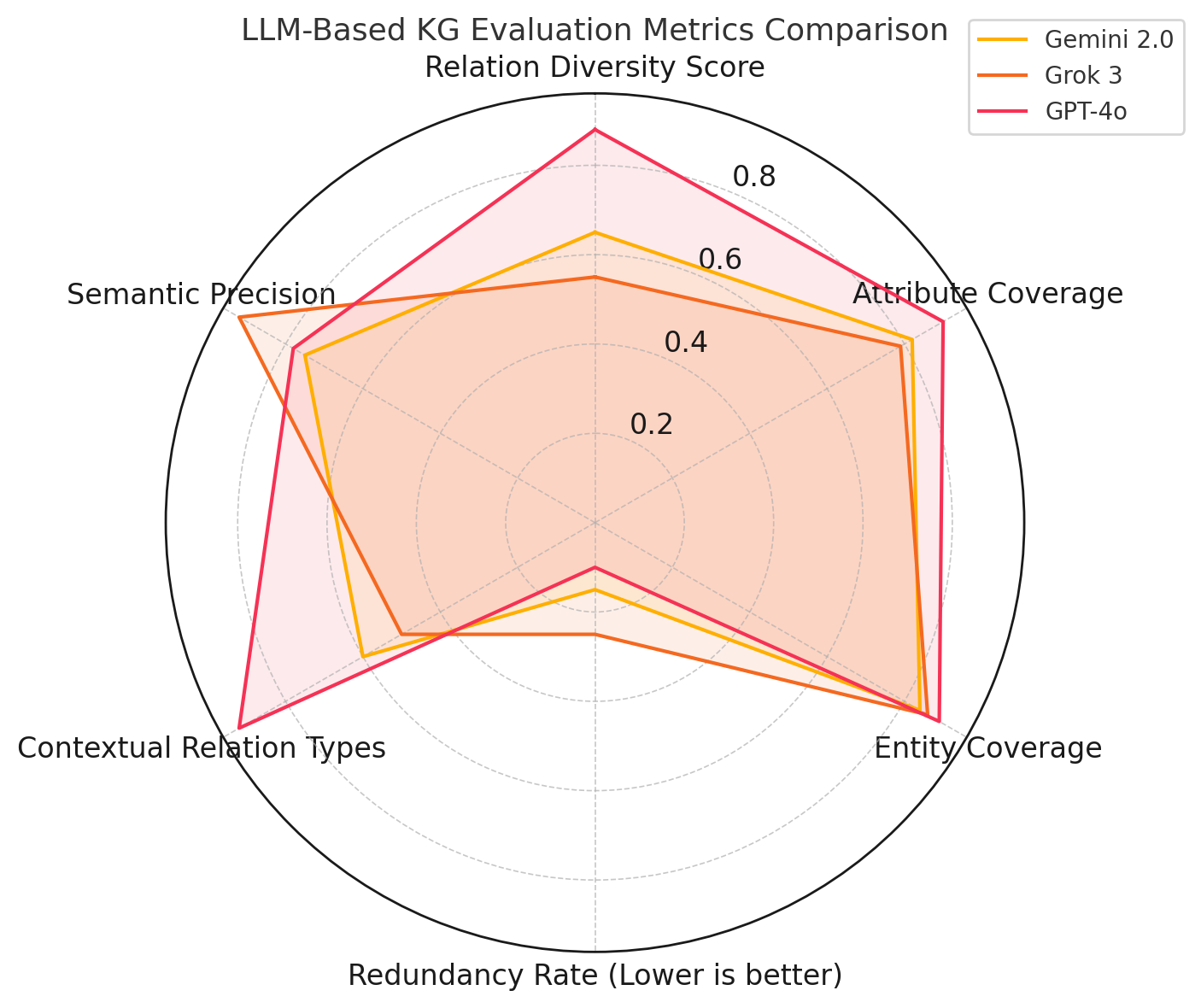}
    \caption{Relation and structural metrics}
\end{subfigure}
\hfill
\begin{subfigure}{0.32\textwidth}
    \includegraphics[width=\linewidth]{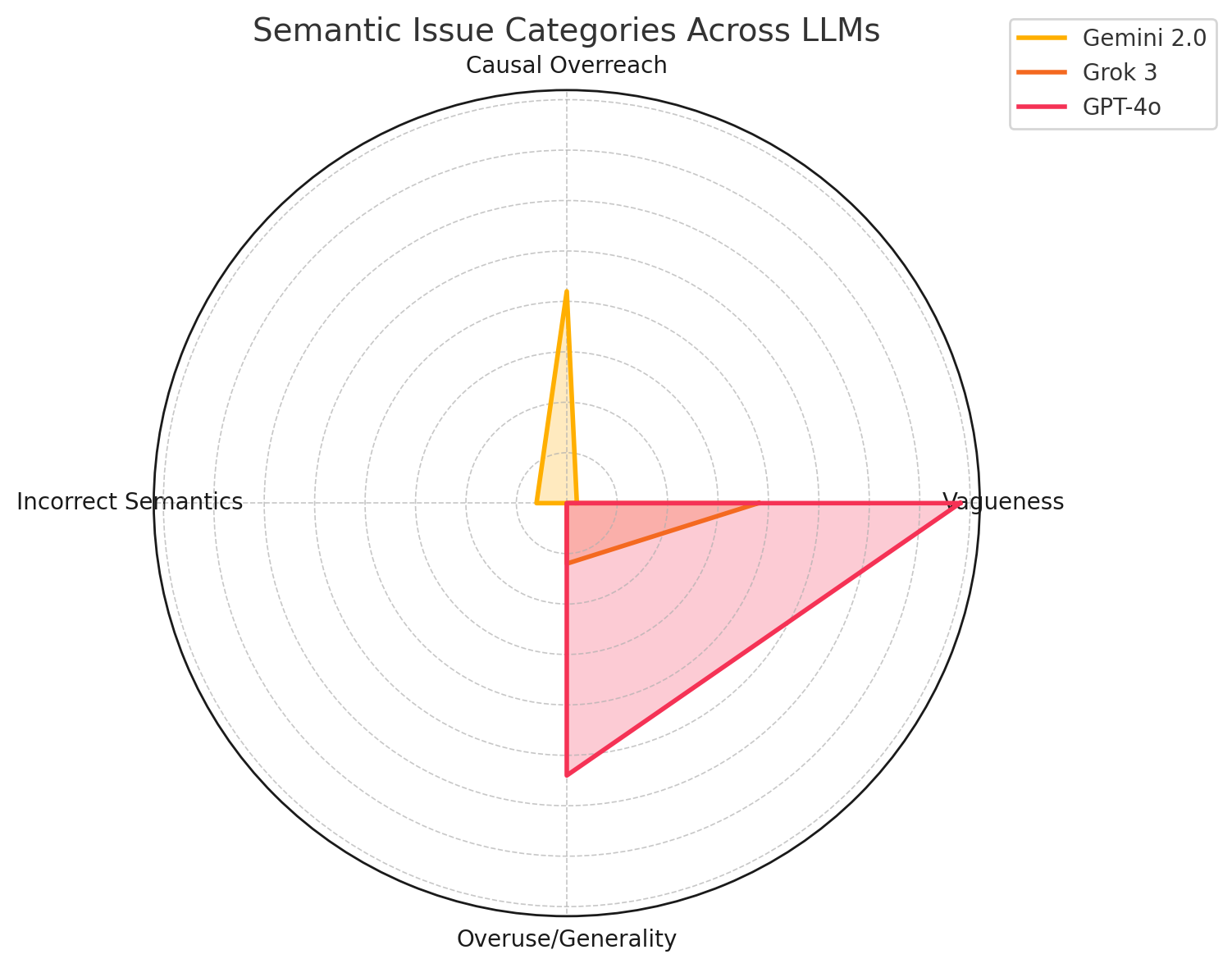}
    \caption{Semantic issue distribution}
\end{subfigure}
\hfill
\begin{subfigure}{0.32\textwidth}
    \includegraphics[width=\linewidth]{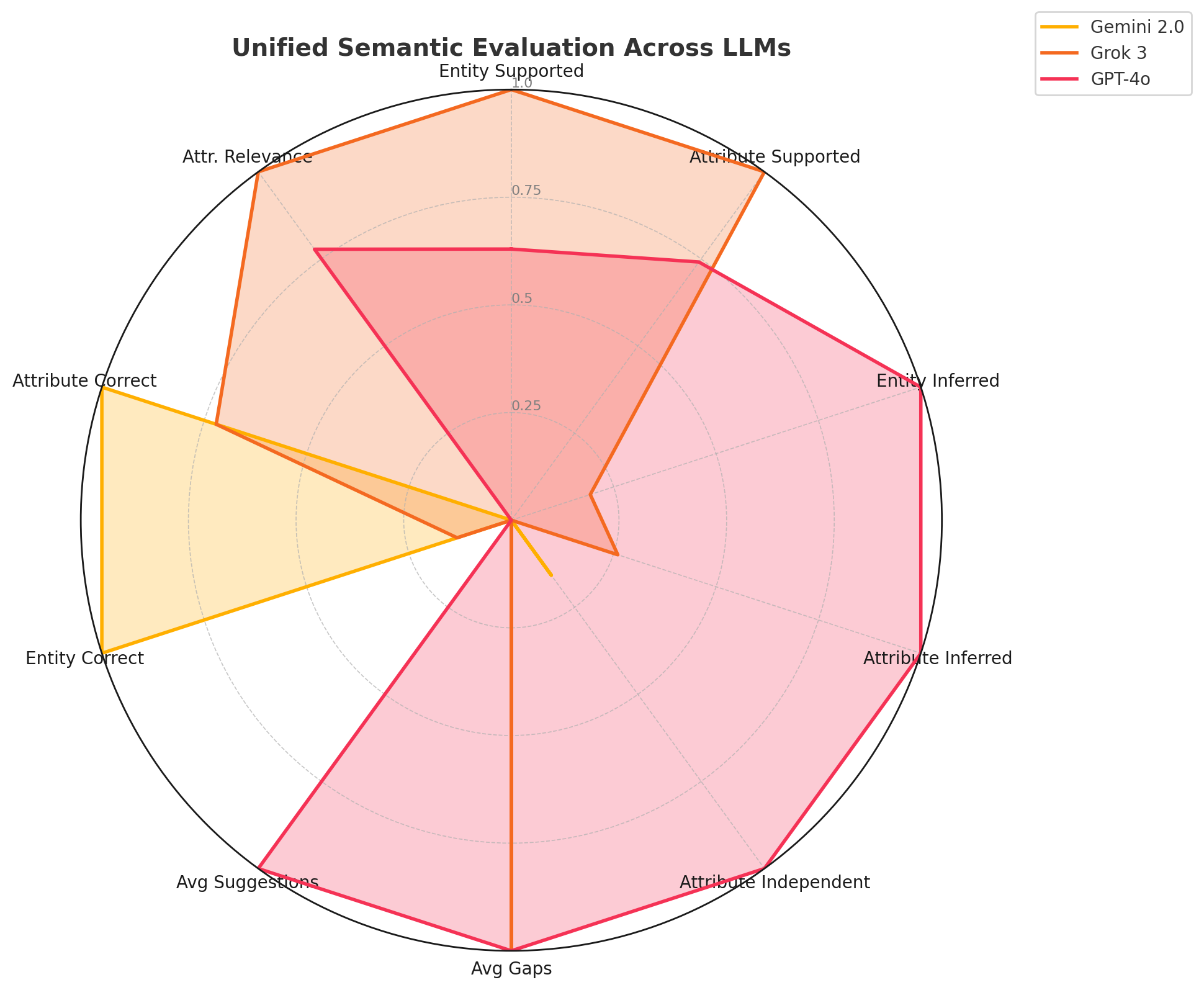}
    \caption{Unified semantic evaluation}
\end{subfigure}
\vspace{-0.1in}
\caption{Radar chart comparison of Gemini 2.0 Flash, Grok 3, and GPT-4o: (a) relation diversity and structural metrics, (b) semantic issue categories, and (c) unified evaluation across correctness, inference, and data support.}
\label{fig:llm_radar_charts_combined}
\end{figure*}

We compared relation types, source coverage, and structural patterns in knowledge graphs generated by \textit{Gemini 2.0 Flash}, \textit{Grok 3}, and \textit{GPT-4o} across 40 oncology reports. Figure~\ref{fig:llm_radar_charts_combined}(a) summarizes key evaluation metrics.

Gemini and Grok focused on core clinical predicates (e.g., \texttt{indicates}, \texttt{treats}, \texttt{influences}), while GPT-4o generated a broader range of descriptive and context-sensitive relations (e.g., \texttt{pertains\_to}, \texttt{assesses}, \texttt{documents}), reflecting its strength in semantic generalization.

All models extracted key clinical entities (\texttt{Patient}, \texttt{Condition}, \texttt{CarePlan}) and biomarkers (\texttt{age}, \texttt{ca\_19\_9}, \texttt{tumor\_grade}). GPT-4o also surfaced nuanced attributes (e.g., \texttt{fertility\_preservation}, \texttt{estrogen\_exposure}), showing sensitivity to subtle contextual cues often missed by the others.

Structurally, each model connected densely around nodes like \texttt{Condition} and \texttt{Treatment}. Gemini emphasized procedural clusters (e.g., \texttt{Assessment}, \texttt{Practitioner}); GPT-4o incorporated diverse concepts (e.g., \texttt{Therapy}, \texttt{FollowUp}) and minimized redundancy. Grok prioritized precision but exhibited more repetition due to conservative extraction.
Figure~\ref{fig:llm_radar_charts_combined}(a) illustrates: GPT-4o excels in relational breadth and abstraction; Grok 3 in semantic precision and redundancy control; and Gemini 2.0 in balanced procedural accuracy.

\begin{table*}[ht!]
\centering
\footnotesize
\setlength{\tabcolsep}{3pt}
\renewcommand{\arraystretch}{1.15}
\caption{Comprehensive evaluation of clinical knowledge graph construction across LLMs with illustrative examples.}
\label{tab:llm_unified_summary_examples}

\begin{tabularx}{\textwidth}{@{}p{3.3cm} p{2.9cm} c c c X@{}}
\toprule
\textbf{Category} & \textbf{Metric} & \textbf{Gemini} & \textbf{Grok} & \textbf{GPT-4o} & \textbf{Example / Description} \\
\midrule

\multirow{3}{*}{\textbf{Data Support}}
& Entity supported (per report) & 21.57 & 24.00 & 23.10 & Entities per patient (\texttt{Patient}, \texttt{Condition}, \texttt{ImagingStudy}). \\
& Attribute supported (per report) & 67.67 & 89.03 & 83.50 & Examples: \texttt{tumor\_size}, \texttt{Ki\mbox{-}67\_Index}, \texttt{weight\_loss}. \\
& Attribute diversity (Unique/Total) & 0.52 & 0.60 & 0.68 & Range of unique attributes, e.g., both \texttt{age} and \texttt{lymph\_node\_count}. \\
\midrule

\multirow{5}{*}{\textbf{Inference/Abstraction}}
& Entity inferred (avg) & 0.12 & 0.28 & 0.95 & Infers \texttt{HER2} implies a \texttt{Biomarker} entity. \\
& Attribute inferred (avg) & 0.70 & 1.35 & 3.20 & Infers \texttt{Prognostic\_Marker} from context. \\
& Attribute independent (no entity) & 0.25 & 0.10 & 1.05 & \texttt{Obesity} extracted without linked \texttt{Patient}. \\
& Implicit inference ratio (\%) & 1.01\% & 1.49\% & 3.92\% & Implicit triple: \texttt{ER\mbox{-}positive} $\rightarrow$ \texttt{responds\_to} $\rightarrow$ \texttt{Tamoxifen}. \\
& Cross-sentence inference (\#) & 1.0 & 1.5 & 3.2 & Combines scattered mentions of \texttt{surgery} and \texttt{blood\_loss}. \\
\midrule

\multirow{3}{*}{\textbf{Gaps/Suggestions}}
& Avg gaps per patient & 1.00 & 2.00 & 2.00 & Missing link: \texttt{Diagnosis} $\rightarrow$ \texttt{associated\_with} $\rightarrow$ \texttt{Biopsy}. \\
& Avg suggestions per patient & 1.00 & 1.00 & 1.50 & Suggests \texttt{liver\_function} for elevated \texttt{bilirubin}. \\
& Gap-fill accuracy (\%) & 55.0\% & 61.0\% & 68.0\% & Fraction of suggestions matching expert-annotated content. \\
\midrule

\multirow{6}{*}{\textbf{Correctness/Relevance}}
& Entity correct (\%) & 99.4\% & 23.3\% & 11.7\% & Correctly identifies \texttt{ImagingStudy} as a clinical entity. \\
& Attribute correct (\%) & 97.5\% & 75.3\% & 18.0\% & Example: \texttt{CA\_19\mbox{-}9} used properly as \texttt{LabResult}. \\
& Attribute relevance (avg) & 0.80 & 0.89 & 0.87 & Clinical utility of \texttt{BMI} and \texttt{Tumor\_Grade}. \\
& Attribute inferred (\%) & 0.95\% & 0.18\% & 2.49\% & Infers \texttt{Smoking\_Status} affects \texttt{Cancer\_Risk}. \\
& Attribute independent (\%) & 0.34\% & 0.00\% & 1.21\% & Isolated attributes not linked to any entity. \\
& Triple validation confidence & 0.96 & 0.91 & 0.88 & Confidence in \texttt{Biopsy} $\rightarrow$ \texttt{confirms} $\rightarrow$ \texttt{TumorType}. \\
\midrule

\multirow{3}{*}{\textbf{Semantic Quality}}
& Hallucination rate (\%) & 0.18\% & 0.04\% & 0.92\% & False triple: \texttt{Vitamin\_D} $\rightarrow$ \texttt{prevents} $\rightarrow$ \texttt{Cancer}. \\
& Redundancy rate (\%) & 2.2\% & 0.7\% & 1.8\% & Relation repetition for the same attribute (e.g., \texttt{indicates}). \\
& SPARQL-compatible triples (\%) & 98.1\% & 99.6\% & 96.2\% & Fraction of triples executable in SPARQL endpoints. \\
\bottomrule
\end{tabularx}
\end{table*}

%\vspace{-0.2in}
\subsubsection*{\bf Semantic Relation Analysis Across LLMs}
To evaluate the semantic integrity of relationships in the clinical knowledge graph, we conducted a comparative analysis of predicate usage by \textit{Gemini 2.0 Flash}, \textit{Grok 3}, and \textit{GPT-4o}. Rather than focusing solely on factual correctness, this analysis emphasizes whether each model employs predicates that are both clinically meaningful and ontologically valid (e.g., \texttt{treats}, \texttt{leads\_to}, \texttt{indicates}).
We identified and grouped common relational errors into four categories: (1) \textit{Vagueness}, involving imprecise predicates such as \texttt{drug\_use vague}; (2) \textit{Causal Overreach}, where unsupported causality is overstated (e.g., \texttt{bp leads\_to comorbidities}); (3) \textit{Incorrect Semantics}, such as the misuse of \texttt{contraindicates} or \texttt{is\_a}; and (4) \textit{Overuse/Generality}, with excessive use of broad predicates like \texttt{indicates} and \texttt{influences}.
Figure~\ref{fig:llm_radar_charts_combined}(b) presents a radar chart summarizing semantic issues. GPT-4o shows more vagueness and generality due to its generative style; Grok 3 minimizes semantic errors through rigorous filtering; Gemini 2.0 exhibits more causal overreach from broader predicate exploration.
These trends suggest that GPT-4o provides relational diversity but may lack precision, Grok 3 emphasizes semantic clarity and restraint, and Gemini 2.0 contributes exploratory richness with some risk of drift. Together, their combined strengths enable predicate refinement and clinical coherence in graph construction.

%\vspace{-0.1in}
\subsubsection*{\bf Clinical Relevance Evaluation Across LLMs}

To assess the clinical relevance of LLM-generated relationships, we compared average entity and attribute relevance across 40 oncology reports. As shown in Table~\ref{tab:llm_unified_summary_examples}, GPT-4o achieved the highest relevance (0.87 for entities and 0.89 for attributes), followed by Grok (0.85 entity, 0.88 attribute), and Gemini (0.80 entity, 0.79 attribute). GPT-4o's strength lies in contextual reasoning; Grok excels in conservative filtering; Gemini trades precision for broader coverage.
These complementary capabilities make the LLM trio well-suited for building clinically grounded knowledge graphs that balance accuracy and discovery, and can be iteratively improved through expert feedback and retrieval-based refinement.

\subsubsection*{\bf Unified Evaluation of LLM Behavior in KG Construction}

We evaluated \textit{Gemini 2.0 Flash}, \textit{Grok 3}, and \textit{GPT-4o} across six dimensions of clinical knowledge graph construction: data support, inference, gap handling, correctness, semantic quality, and SPARQL compatibility. Table~\ref{tab:llm_unified_summary_examples} provides a detailed comparison with clinical examples.
\textit{Gemini 2.0} excels in precision, achieving the highest entity (99.4\%) and attribute correctness (97.5\%), and high SPARQL compliance (98.1\%), though with limited inference and attribute diversity due to its schema-constrained approach.

\textit{Grok 3} offers strong semantic rigor and moderate inference, with high attribute relevance (0.89), minimal hallucination (0.04\%), and low redundancy—ideal for conservative, high-precision graph construction.
\textit{GPT-4o} leads in contextual inference and semantic enrichment, extracting diverse attributes 
(e.g., \texttt{estrogen\_exposure}) and achieving the highest gap fill accuracy (68.0\%). While hallucination is slightly higher, its strength lies in uncovering implicit relations and improving graph expressiveness.
In the absence of ground truth, our multi-agent ensemble ensures robustness by combining Gemini’s structured extraction, Grok’s semantic validation, and GPT-4o’s contextual enrichment. This yields ontology-aligned, clinically meaningful graphs suitable for decision support and advanced analytics.

\section{Limitations and Conclusion}

Despite its strengths, the framework has several limitations. It is not yet integrated with live clinical decision support systems (CDSS), limiting real-world deployment. It currently operates solely on unstructured clinical text, omitting other critical modalities such as imaging (CT, MRI), biosignals (ECG, EMG), and structured EHR data. Our evaluation, based on 40 oncology reports (PDAC and BRCA) from the CORAL dataset, also limits generalizability across broader clinical settings. While model consensus and ontology alignment provide a form of weak supervision, expert-in-the-loop validation remains essential for resolving ambiguous or domain-specific triples.

We introduce the first framework to construct and evaluate clinical knowledge graphs directly from free-text reports using multi-LLM consensus and a Retrieval-Augmented Generation (RAG) strategy. Our pipeline combines schema-guided EAV extraction via \textit{Gemini 2.0 Flash}, semantic refinement by \textit{GPT-4o} and \textit{Grok 3}, and ontology-aligned encoding using SNOMED CT, RxNorm, LOINC, GO, and ICD in RDF/RDFS/OWL with SWRL rules.

This KG-RAG approach supports continuous refinement, enabling iterative improvement of graph quality through self-supervised validation and expert feedback. By retrieving, validating, and grounding extracted triples over time, the system evolves dynamically, bridging static KG generation with adaptive knowledge refinement. Gemini anchors high-precision extraction, Grok filters hallucinations and enforces semantic rigor, and GPT-4o contributes contextual depth and abstraction, yielding clinically relevant, SPARQL-compatible graphs.

Future directions include real-time CDSS integration, multimodal fusion, large-scale validation across institutions, and enriched temporal and causal modeling. Ultimately, our framework provides a scalable and explainable foundation for transforming unstructured narratives into trustworthy, ontology-grounded clinical knowledge graphs.

\bibliographystyle{plain}
\bibliography{References}

\begin{thebibliography}{10}

\bibitem{ayaz2021fast}
Muhammad Ayaz, Muhammad~F Pasha, Mohammed~Y Alzahrani, Rahmat Budiarto, and Deris Stiawan.
\newblock The fast health interoperability resources (fhir) standard: systematic literature review of implementations, applications, challenges and opportunities.
\newblock {\em JMIR Medical Informatics}, 9(7):e21929, 2021.

\bibitem{cao2023hallucination}
Shuyang Cao, Lu~Wang, and Dragomir Radev.
\newblock A survey on hallucination in natural language generation.
\newblock {\em ACM Computing Surveys (CSUR)}, 55(12):1--38, 2023.

\bibitem{gemini2_2024}
Google DeepMind.
\newblock Gemini 2.0 technical report.
\newblock \url{https://deepmind.google/technologies/gemini/}, 2024.
\newblock Accessed May 2025.

\bibitem{gubanov2024cancerkg}
Michael Gubanov, Anna Pyayt, and Aleksandra Karolak.
\newblock Cancerkg. org-a web-scale, interactive, verifiable knowledge graph-llm hybrid for assisting with optimal cancer treatment and care.
\newblock In {\em Proceedings of the 33rd ACM International Conference on Information and Knowledge Management}, pages 4497--4505, 2024.

\bibitem{kandibedala2023covidkg}
Bhimesh Kandibedala, Anna Pyayt, Nickolas Piraino, Chris Caballero, and Michael Gubanov.
\newblock Covidkg.org--a web-scale covid-19 interactive, trustworthy knowledge graph, constructed and interrogated for bias using deep-learning.
\newblock In {\em Proceedings of the International Conference on Extending Database Technology (EDBT)}, 2023.

\bibitem{khan2020weblens}
Rituparna Khan and Michael Gubanov.
\newblock Weblens: Towards web-scale data integration, training the models.
\newblock In {\em 2020 IEEE International Conference on Big Data (Big Data)}, pages 5727--5729. IEEE, 2020.

\bibitem{lin2022truthfulqa}
Stephanie Lin, Jacob Hilton, and Owain Evans.
\newblock Truthfulqa: Measuring how models mimic human falsehoods.
\newblock {\em Proceedings of the 60th Annual Meeting of the Association for Computational Linguistics (ACL)}, pages 3214--3252, 2022.

\bibitem{liu2023kala}
Shizhe Liu, Chong Zheng, Yuxian Wang, Zhengyan Zhang, Zhiyuan Liu, and Maosong Sun.
\newblock Kala: Knowledge-augmented language model pretraining.
\newblock {\em arXiv preprint arXiv:2306.11644}, 2023.

\bibitem{liu2022knowgl}
Zhiyuan Liu, Zhengyan Zhang, Yuxian Wang, Yantao Shen, Zhicheng Liu, Jiliang Tang, and Maosong Sun.
\newblock Knowgl: Knowledge generation and linking from text.
\newblock In {\em Proceedings of the 60th Annual Meeting of the Association for Computational Linguistics (ACL)}, pages 4364--4378, 2022.

\bibitem{openai2024gpt4o}
{OpenAI}.
\newblock Gpt-4o.
\newblock \url{https://openai.com/index/gpt-4o}, 2024.
\newblock Accessed: 2025-05-30.

\bibitem{rajpurkar2023evaluating}
Pranav Rajpurkar, Jeremy Irvin, Haoran Zhang, Brandon Yang, Harsh Mehta, Katie Shpanskaya, Bobak~J Wang, Riley Jones, Kevin~H Yu, and Matthew~P Lungren.
\newblock Evaluating the factual consistency of large language models in medical applications.
\newblock {\em NPJ Digital Medicine}, 6(1):1--13, 2023.

\bibitem{loinc2022}
{Regenstrief Institute}.
\newblock Loinc: Logical observation identifiers names and codes.
\newblock \url{https://loinc.org}, 2022.
\newblock Accessed May 2025.

\bibitem{shinn2023reflexion}
Noah Shinn, Yuchen Tang, Harsha Nori, Vidhi Agarwal, Jianfeng Gao, and Xinyun Chen.
\newblock Reflexion: Language agents with verbal reinforcement learning.
\newblock {\em arXiv preprint arXiv:2303.11366}, 2023.

\bibitem{simmons2017hybridjson}
Mark Simmons, Daniel Armstrong, Dylan Soderman, and Michael Gubanov.
\newblock Hybrid.json: High-velocity parallel in-memory polystore json ingest.
\newblock In {\em 2017 IEEE International Conference on Big Data (Big Data)}, pages 2741--2750. IEEE, 2017.

\bibitem{snomed2022}
{SNOMED International}.
\newblock Snomed ct: Systematized nomenclature of medicine -- clinical terms.
\newblock \url{https://www.snomed.org/snomed-ct}, 2022.
\newblock Accessed May 2025.

\bibitem{sushil2024coral}
Madhumita Sushil, Vanessa~E Kennedy, Divneet Mandair, Brenda~Y Miao, Travis Zack, and Atul~J Butte.
\newblock Coral: expert-curated oncology reports to advance language model inference.
\newblock {\em NEJM AI}, 1(4):AIdbp2300110, 2024.

\bibitem{rxnorm2022}
{U.S. National Library of Medicine}.
\newblock Rxnorm: Normalized names for clinical drugs.
\newblock \url{https://www.nlm.nih.gov/research/umls/rxnorm/}, 2022.
\newblock Accessed May 2025.

\bibitem{grok2024}
xAI.
\newblock Grok: Ai by xai (elon musk's company).
\newblock \url{https://x.ai}, 2024.
\newblock Accessed May 2025.

\bibitem{xue2024clr2g}
Hongchen Xue, Qingzhi Ma, Guanfeng Liu, Jianfeng Qu, Yuanjun Liu, and An~Liu.
\newblock Clr2g: Cross modal contrastive learning on radiology report generation.
\newblock In {\em Proceedings of the 33rd ACM International Conference on Information and Knowledge Management}, pages 2742--2752, 2024.

\bibitem{yang2023gatortronrag}
Xi~Yang, Yichi Zhang, Yuxing Si, Yuhao Zhang, Meng Jiang, Fei Wang, and Hua Xu.
\newblock Gatortron-rag: A retrieval-augmented generation framework for biomedical question answering.
\newblock {\em arXiv preprint arXiv:2311.00964}, 2023.

\bibitem{zhang2024knowgpt}
Rui Zhang, Jack Syu, Edward Hu, Kai-Wei Chang, and Hao Tan.
\newblock Knowgpt: Benchmarking and improving knowledge grounding of large language models.
\newblock {\em arXiv preprint arXiv:2402.08600}, 2024.

\bibitem{zhao2025medrag}
Xuejiao Zhao, Siyan Liu, Su-Yin Yang, and Chunyan Miao.
\newblock Medrag: Enhancing retrieval-augmented generation with knowledge graph-elicited reasoning for healthcare copilot.
\newblock In {\em Proceedings of the ACM on Web Conference 2025}, pages 4442--4457, 2025.

\bibitem{zhu2024emerge}
Yinghao Zhu, Changyu Ren, Zixiang Wang, Xiaochen Zheng, Shiyun Xie, Junlan Feng, Xi~Zhu, Zhoujun Li, Liantao Ma, and Chengwei Pan.
\newblock Emerge: Enhancing multimodal electronic health records predictive modeling with retrieval-augmented generation.
\newblock In {\em Proceedings of the 33rd ACM International Conference on Information and Knowledge Management}, pages 3549--3559, 2024.

\end{thebibliography}

\end{document}